# Leveraging Long-Context Large Language Models for Multi-Document Understanding and Summarization in Enterprise Applications

Aditi Godbole[1][0009-0004-8407-9541] and Jabin Geevarghese George[2][0009-0005-8716-0385] and Dr. Smita Shandilya[3]

[1] IEEE Senior Member, Seattle, USA
[2] TCS, New Jersey, USA
[3] Institute of Advance Computing (Spl. Data Science), SAGE University, Indore, India

`aditi.godbole@ieee.org`

**Abstract.** The rapid increase in unstructured data across various fields has made multi-document comprehension and summarization a critical task. Traditional approaches often fail to capture relevant context, maintain logical consistency, and extract essential information from lengthy documents. This paper explores the use of Long-context Large Language Models (LLMs) for multi-document summarization, demonstrating their exceptional capacity to grasp extensive connections, provide cohesive summaries, and adapt to various industry domains and integration with enterprise applications/systems. The paper discusses the workflow of multi-document summarization for effectively deploying long-context LLMs, supported by case studies in legal applications, enterprise functions such as HR, finance, and sourcing, as well as in the medical and news domains. These case studies show notable enhancements in both efficiency and accuracy. Technical obstacles, such as dataset diversity, model scalability, and ethical considerations like bias mitigation and factual accuracy, are carefully analyzed. Prospective research avenues are suggested to augment the functionalities and applications of long-context LLMs, establishing them as pivotal tools for transforming information processing across diverse sectors and enterprise applications.

## 1 Introduction

In a multinational corporation, a team of analysts faces the daunting task of summarizing thousands of documents spanning financial reports, market analyses, and internal communications to inform a critical strategic decision. This scenario illustrates the challenge of multi-document summarization in enterprise settings, where the volume and diversity of information can overwhelm traditional analysis methods.

The exponential growth of unstructured text data across various sectors has made document summarization a critical task [1]. Multi-document summarization presents unique challenges due to the need for synthesizing information from diverse sources, which may contain redundant, complementary, or contradictory information across documents [4]. Variations in writing style and level of detail add complexity to the task. Determining the relevance and importance of information from each source is crucial for creating a coherent and comprehensive summary [5].

Traditional document summarization techniques often struggle with redundancy, inconsistency, lack of context understanding, scalability issues for multiple document summarization tasks, inability to capture cross-document relationships, difficulty handling diverse formats, and lack of domain adaptability [6, 7, 8]. These limitations highlight the need for more advanced approaches to multi-document summarization.

This paper addresses the following research question: How can long-context Large Language Models (LLMs) be leveraged to improve multi-document understanding and summarization in enterprise applications?

We investigate the use of Long-context LLMs for multi-document summarization, demonstrating their exceptional capacity to grasp extensive connections, provide cohesive summaries, and adapt to various industry domains and integration with enterprise applications/systems [14]. The paper discusses the workflow of multi-document summarization for effectively adopting long-context LLMs, supported by case studies in legal applications, enterprise functions such as HR, finance, and sourcing, as well as in the medical and news domains [46, 49, 53].

By exploring the potential of Long-context LLMs in multi-document summarization, we aim to address the limitations of traditional methods and provide a more efficient and accurate approach to processing large volumes of unstructured data [13, 18]. This research has significant implications for improving information processing across diverse sectors and enterprise applications.

## 2    Traditional Document Summarization Techniques and Challenges

### 2.1    Traditional Methods of Document Summarization

Document summarization involves creating a concise version of a document while retaining its key ideas and essential information. The process typically includes several key steps: text analysis to understand document structure and themes, information extraction to identify relevant content, content selection to choose the most important elements, summary generation, and quality evaluation [1].

Traditional approaches to summarization fall into two main categories: extractive and abstractive methods. Extractive summarization selects and arranges existing sentences or phrases from the source document, while abstractive summarization generates new text to capture the core concepts [2].

Extractive techniques often rely on statistical measures of word frequency, sentence position, and other surface-level features to identify important content [1]. More advanced extractive methods use graph-based algorithms to model relationships between sentences and identify central ideas [7]. Machine learning approaches have also been applied to learn patterns for selecting relevant sentences [2].

Abstractive summarization is generally considered more challenging, as it requires language generation capabilities. Early abstractive methods often used templated approaches, filling in extracted information into predefined sentence structures [3]. With the rise of neural networks, sequence-to-sequence models became a popular approach for abstractive summarization [18].

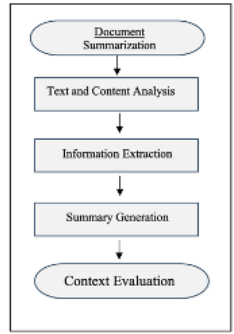

**Fig. 1.** Traditional Document Summarization Techniques

## 2.2   Challenges in Multi-Document Summarization

While these traditional techniques can be effective for single documents, they face significant challenges when applied to multi-document summarization:

Redundancy: Information is often repeated across multiple source documents, leading to repetitive summaries if not properly managed [4].

Coherence: Extracting sentences from different documents can result in disjointed summaries lacking logical flow [5].

Context preservation: It is challenging to maintain the broader context when combining information from diverse sources [6], leading to summaries that lack focus or miss key points[11].

Scalability: Processing and synthesizing information from many documents increases computational complexity [10]. This challenge persists even with advanced models, as handling very long or numerous documents remains computationally intensive [14].

Cross-document relationships: Capturing connections and contradictions between documents is difficult for methods focused on individual sentences or documents [7].

Domain adaptation: Traditional methods can be limited in their flexibility and adaptability to handle diverse document structures and writing styles [9]. Techniques optimized for one type of document may not generalize well to other domains or writing styles [12]. This challenge extends to ensuring models can handle varied document formats, styles, and topics [59].

Factual consistency: Ensuring that generated summaries remain faithful to the source material is crucial, especially for abstractive methods [66]. This challenge is particularly pronounced in abstractive summarization, where preventing "hallucinations" or factual errors remains an active area of research [65].

Ethical considerations: As summarization techniques become more advanced, mitigating biases present in training data and ensuring fairness in generated summaries becomes increasingly important [63].

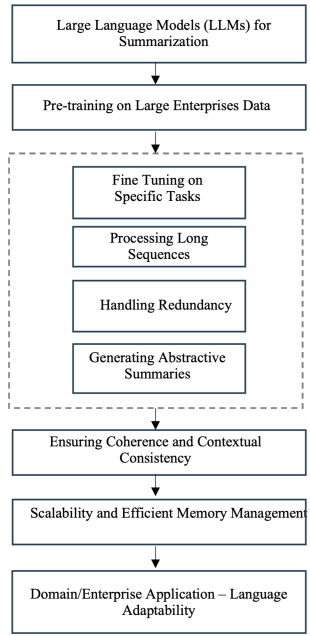

**Fig. 2.** LLM Workflow for Document Summarization

### 2.3   Large Language Models for Document Summarization

Large Language Models (LLMs) have emerged as a promising solution to address many of these challenges in multi-document summarization. These models, based on transformer architectures [13], can process and understand much longer sequences of text than previous approaches. This allows them to capture broader context and relationships across multiple documents [14].

Key advantages of LLMs for multi-document summarization include:

Contextual understanding: Self-attention mechanisms allow LLMs to model long-range dependencies and capture document-level context [15].

Redundancy handling: LLMs can identify and consolidate repeated information across sources [16].

coherent and contextually consistent summaries: the hierarchical structure of LLMs allows them to model the inherent structure of documents, which can create more coherent and contextually consistent summaries [17].

Abstractive capabilities: Through pre-training on vast corpora, LLMs develop strong language generation abilities, enabling more natural abstractive summaries [18].

Scalability: Techniques like sparse attention [20] allow LLMs to efficiently process very large inputs.

Transfer learning: LLMs can be fine-tuned for specific summarization tasks while leveraging general language understanding [19, 21].

### 2.4 Domain-Specific Applications

Researchers have begun exploring the application of LLMs to multi-document summarization in various specialized domains:

Legal: LLMs show promise in summarizing complex legal documents, extracting key arguments and precedents [46]. This can significantly improve efficiency for legal professionals reviewing large case files [47].

Medical: In the medical field, LLMs are being used to synthesize findings from multiple research papers, supporting evidence-based practice [49]. They can also summarize patient records and clinical conversations to aid healthcare providers [52].

News: LLMs enable the aggregation of multiple news sources to provide comprehensive event summaries, potentially reducing bias by incorporating diverse viewpoints [53, 55].

These domain-specific applications highlight the adaptability of LLMs to different types of documents and summarization needs. However, they also underscore the importance of addressing potential biases and ensuring factual accuracy, especially in sensitive domains [62, 65].

Long context understanding within large language models is crucial for document summarization tasks requiring a deep understanding of language semantics, coherence, and context. Long Context LLMs offer a powerful approach to overcome the limitations of traditional methods in multi-document summarization. By capturing long-range dependencies, handling redundancy, generating coherent abstracts, and scaling to large document collections, LLMs have the potential to revolutionize the field of multi-document summarization and enable the creation of more accurate, informative, and user-friendly summaries. In this paper, we describe the application of Long-context LLMs through various case studies.

## 3 Methodology

A typical multi-document summarization task follows the steps described below.

Model Selection: A Long context LLM such as GPT-4 [22] or Claude 2.1 [23] can be chosen for multi-document summarization task. The final model is selected for its ability to capture long-range dependencies and handle large input sequences and generate coherent and fluent summaries [24]. An LLM's pre-training on diverse datasets allows for effective transfer learning to the multi-document summarization task.

Data preparation: In data preparation step, multiple documents relevant for the summarization task having balanced representation of multiple sources such as news articles, academic papers, and reports [25] are collected to create a comprehensive dataset. The documents are filtered and to remove duplicates, irrelevant content and noise to ensure data quality. For data preprocessing the text is normalized, tokenized, and formatted for input to the LLMs.

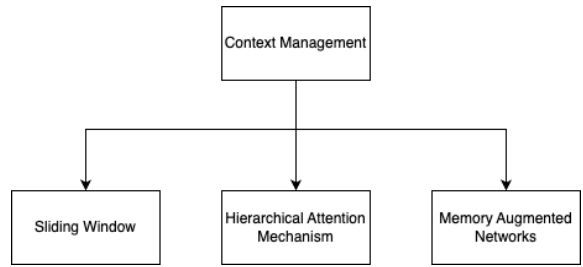

**Fig. 3.** Context Management Techniques

Context Management: In order, to reduce the complexity of managing longer context we have selected Long-context LLMs, however, even explicit Long-context LLMs may suffer from losing the context when relevant context appears in the middle of a very long text [26], hence managing long contexts is critical for effective summarization. Sliding window, hierarchical attention and use of memory- augmented networks are some of the strategies that are effective. Sliding Window involves dividing the text into overlapping segments to ensure the model captures context across sections. [27]. In Hierarchical Attention Mechanisms layers of attention are implemented to focus on different levels of text (e.g., paragraphs, sentences) for better context retention. [28]. Memory-augmented networks, such as the Differentiable Neural Computer (DNC) [29] and the Memory Attention Network (MAN) [30], are integrated with the LLMs to enable dynamic context retrieval and information storage during the summarization process where it employs external memory structures to dynamically store and retrieve context information, enhancing the model's ability to reference long-range dependencies.

Information Extraction: This is a crucial step in multi-document summarization, that involves identifying and organizing key pieces of information across multiple documents to generate coherent and comprehensive summaries. Named Entity Recognition (NER), Relation Extraction (RE) and coreference resolution are the techniques most used in Information Extraction step where NER helps to identify and classify entities such as people, organizations, locations, dates, and other important terms within the text and RE assists to identify relationships between entities, such as who did what to whom, when, and where [31]. Coreference Resolution aims to identify, and link mentions of the same entity across different parts of the text or across multiple documents [32]. By applying NER to the input documents, LLM can recognize and extract key entities that are central to the information being summarized [33]. The extracted entities serve as anchors for linking and integrating information across multiple documents. LLMs can utilize RE techniques to uncover the connections and interactions among the extracted entities, providing a deeper understanding of the information structure [34]. Extracting relations helps in establishing the context and coherence of the information being summarized. By applying Coreference Resolution, LLMs can establish connections between related entities and events, even if they are referred to using different expressions [35]. This helps in creating a unified representation of the information and avoids redundancy in the generated summary.

Information Integration: In this step, the extracted entities and relations are used to construct knowledge graphs that capture the interconnections and hierarchical structure of the information [36]. These knowledge graphs aid in maintaining coherence and consistency during summary generation.

Summary Generation: In this steps extractive and abstractive summarization approaches are combined by employing the hybrid approach [37]. Extractive methods are used to identify the most salient sentences, while abstractive methods generate concise and fluent summaries. This approach ensures coherence, consistency, and relevance in the generated summaries: To ensure the quality of the generated summaries, techniques such as coherence modeling [38], consistency checking [39], and relevance scoring [40] are incorporated. These techniques help in maintaining the logical flow, factual accuracy, and pertinence of the summaries to the input documents.

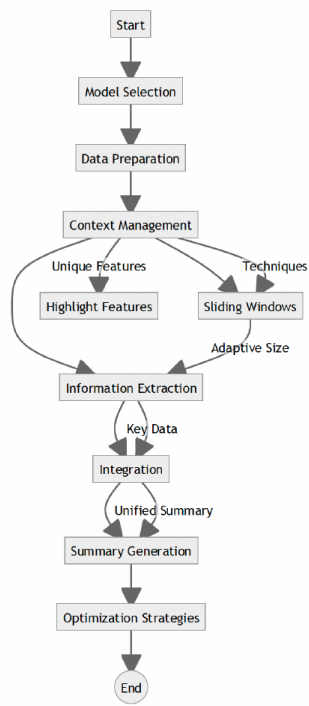

**Fig. 4.** Workflow Analysis for Leveraging LLMs in Multi-Document Summarization

Optimization Strategies: The pre-trained LLMs are fine-tuned on multi-document summarization datasets using techniques such as transfer learning [41] and domain adaptation 42]. This allows the models to specialize in the specific task and domain of interest. To optimize the performance of the LLMs, techniques such as model compression [43], knowledge distillation [44], and quantization [45] are employed. These techniques help in reducing the computational overhead while maintaining the accuracy of the generated summaries.

## 4      Applications and Case Studies

In this section we discuss the application and case studies for using long-context large language models (LLMs) for multi-document summarization in the legal domain, medical field, news industry and enterprise application use cases.

### 4.1     Legal Domain

Problem statement: A legal firm needs to summarize a large collection of legal documents related to a complex corporate litigation case. The documents include court filings, depositions, contracts, and legal precedents spanning thousands of pages.

Using a long-context LLM, the legal team can process and summarize the vast amount of legal information efficiently. The LLM identifies key legal entities, extracts relevant clauses and arguments, and generates a concise summary that captures the essential points of the case [46].

This solution will enhance the efficiency by automating the time-consuming task of summarizing lengthy legal documents, allowing legal professionals to allocate their time and resources more efficiently [47] LLMs can identify and highlight crucial legal details that might be overlooked by human readers, reducing the risk of errors or omissions there by resulting in higher accuracy. The summarized legal information is more easily accessible and understandable for legal professionals, enabling them to quickly grasp the essential points without reading through voluminous documents [40]. Using Long-context LLM for multi document summarization in the legal domain also allows for improved decision making as the summaries generated by LLMs provide lawyers with a comprehensive overview of the case, enabling them to make informed decisions and develop effective legal strategies [48].

### 4.2     Medical Field

Problem statement: Researchers are conducting a systematic review of medical literature on a specific disease. They need to summarize findings from hundreds of scientific papers to identify trends, compare treatment options, and guide clinical decision-making.

Employing a long-context LLM, the researchers can efficiently process the large corpus of medical literature. The LLM extracts key information such as study objectives, methodologies, results, and conclusions, generating a structured summary of the reviewed papers [49].

Using a long context LLM can allow healthcare researchers and providers stay current with the latest research findings by summarizing relevant medical literature [50]. Summaries generated by LLMs provide a consolidated view of scientific evidence, supporting healthcare researchers and providers in making evidence-based decisions [51]. Healthcare researchers and providers can quickly access the summarized research findings, enabling them to stay updated with the latest medical knowledge without spending extensive time reading full-text articles [52]. by quickly accessing summarized medical information, healthcare providers can make more informed diagnoses, select appropriate treatments, and enhance overall patient care [52].

### 4.3 News Industry

Problem statement: A news organization wants to create comprehensive summaries of news events by aggregating articles from multiple sources. The goal is to provide readers with a balanced and informative overview of the event.

Utilizing a long-context LLM, the news organization can process articles from various news outlets, identifying key facts, opinions, and perspectives. The LLM generates a coherent summary that combines information from multiple sources, presenting a comprehensive narrative of the event [53]. Multi-document summaries generated by LLMs can provide readers with a broad understanding of news events, reducing the need to read multiple articles [54]. LLMs can incorporate diverse viewpoints from different sources, promoting balanced and unbiased reporting [55]. News organizations can use LLMs to automate the process of summarizing news events, saving time and resources while maintaining high-quality journalism [56].

### 4.4 Enterprise Applications

Problem statement: A large financial services bank needs to summarize a large collection of documents related to various functions such as HR, finance, sourcing, compliance, and audit to streamline operations and improve decision-making processes.

Utilizing long-context LLMs in enterprise functions allows for the condensation of different document types, resulting in improved efficiency, decision-making, and accuracy throughout the organization. Within HR functions, extensive LLMs are utilized to condense employee performance reviews, training documents, and compliance reports. This offers HR managers a thorough understanding of employee performance and compliance status, eliminating the need to sift through lengthy documents. Within finance functions, LLMs provide valuable support to financial analysts by condensing financial reports, investment analyses, and market research. Analysts can efficiently access crucial insights and make well-informed decisions using summarized financial data.

When it comes to sourcing functions, LLMs with extensive knowledge provide concise summaries of supplier contracts, procurement reports, and market analyses. This assists sourcing managers in making strategic decisions by offering them succinct summaries of procurement-related documents. LLMs play a crucial role in summarizing compliance reports, audit findings, and regulatory documents for compliance and audit functions. This guarantees strict compliance with legal standards and enhances the efficiency of audits by emphasizing important compliance issues and audit findings.

The advantages for enterprise operations can be seen through increased efficiency as it streamlines the process of summarizing lengthy documents, allowing enterprise professionals to save their valuable time.

It offers succinct and pertinent summaries that enable managers to swiftly make well-informed decisions.

These case studies demonstrate the practical applications and benefits of using long-context LLMs for multi-document summarization in various domains. By leveraging the power of LLMs, professionals in the legal domain, medical field, and news industry and enterprises irrespective of their size can efficiently process and summarize large volumes of information, leading to improved decision-making, knowledge sharing, and public understanding.

## 5    Challenges and Considerations

The application of long-context LLMs to multi-document summarization, while promising, presents several significant challenges that warrant careful consideration. These challenges span technical, ethical, and practical dimensions, each requiring innovative solutions to ensure the effective and responsible use of these powerful models [56]. Technical Consideration

### 5.1    Technical Considerations

Multi-document summarization often involves processing datasets with diverse formats, styles, and topics, which can pose challenges for LLMs [59]. LLMs need to effectively handle the complexity and heterogeneity of the input data, including variations in document length, structure, and quality [60] Addressing these challenges requires robust data preprocessing techniques, such as document segmentation, noise reduction, and format normalization [61]. Processing large volumes of documents for multi-document summarization can be computationally intensive, requiring significant resources and time [14]. Techniques such as model compression, distributed computing, and hardware acceleration can be employed to improve the scalability and efficiency of LLMs for multi-document summarization [56].

### 5.2    Ethical Considerations

The potential for bias in LLMs raises significant ethical concerns. These models may inadvertently perpetuate or amplify existing biases present in their training data, including gender, racial, cultural, or ideological biases [62, 63]. The societal impact of these biases in summarization tasks cannot be overstated, potentially influencing public opinion or decision-making processes in fields such as journalism or policy-making [64].

Privacy concerns are particularly critical when summarizing sensitive documents, especially in domains like healthcare or legal services where confidentiality is of utmost importance[65].

### 5.3 Ensuring factual accuracy and reliability of summaries

Generating summaries that are factually accurate and reliable is crucial, especially in domains such as news, legal, and medical [65]. LLMs may sometimes generate summaries that contain inconsistencies, hallucinations, or errors, compromising the reliability of the output [66]. This challenge is particularly pronounced in abstractive summarization, where models generate new text rather than extracting existing sentences.

LLM's performance can degrade when relevant information is positioned in the middle of long input context [68], leading to inconsistent performance, incomplete or inaccurate summaries, and reduced coherence and readability. This issue highlights the trade-offs between abstractive and extractive summarization in terms of factual accuracy and coherence [67].

These challenges can be addressed using techniques such as fact-checking, source attribution, and uncertainty quantification to improve the factual accuracy and reliability of the generated summaries [67]. Rigorous evaluation and analysis of the generated summaries will be crucial to identify and address any performance issues related to the positioning of relevant information in long contexts.

### 5.4 Emerging Challenges

As the field of multi-document summarization evolves, new challenges are emerging. Multimodal summarization, involving documents with text, images, and other media types, requires models to integrate information across different modalities coherently. Multilingual and cross-lingual summarization present unique challenges in an increasingly globalized world [68].

Developing comprehensive evaluation metrics for summarization quality remains an open challenge [69]. Current metrics often fail to capture nuanced aspects of summary quality, such as coherence, relevance, and faithfulness to the original documents.

The "black box" nature of LLMs raises concerns about explainability and transparency in the summarization process [70]. As these models are increasingly used in critical applications, there is a growing need for methods to interpret and explain their decision-making processes.

Addressing these challenges will require ongoing research and collaboration across disciplines. As we continue to advance the capabilities of long-context LLMs for multi-document summarization, it is crucial to remain mindful of these considerations to ensure the development of effective, ethical, and reliable summarization systems.

## 6     Conclusion And Future direction

Long-context large language models (LLMs) have emerged as a powerful tool for multi-document understanding and summarization. This study has demonstrated their significant advantages over traditional methods, including the ability to capture long-range dependencies, handle redundancy effectively, and generate coherent summaries. The application of long-context LLMs for multi-document summarization shows potential to revolutionize information processing and knowledge dissemination across various domains.

Our research, supported by case studies in legal applications, enterprise functions, medical, and news domains, highlights several key findings. Long-context LLMs demonstrate superior ability in synthesizing information across multiple documents and show effectiveness in adapting to various fields, producing relevant and coherent summaries. These models excel at consolidating information from diverse sources, resulting in comprehensive and concise summaries. Importantly, the use of these models led to tangible improvements in decision-making, knowledge sharing, and public understanding of complex information.

However, our study also identified important challenges that require further research. Technical challenges include handling diverse and complex datasets, mitigating performance degradation when crucial information is positioned in the middle of long contexts, and scaling models efficiently for very large document collections. Ethical considerations are equally important, encompassing the need to address biases present in training data, ensure factual accuracy and reliability of generated summaries, and maintain privacy and confidentiality, especially in sensitive domains.

Future studies in multi-document summarization using long-context LLMs should prioritize several important factors. Researchers should focus on incorporating domain-specific knowledge bases and ontologies to improve semantic comprehension and coherence of generated summaries. Applying long-context LLMs to various domains, including financial reports, scientific literature, and social media, will enable the creation of customized summaries for specific industry requirements. Further investigation is needed to uncover the possibilities of LLMs in cross-lingual and multilingual summarization, potentially enhancing information accessibility across linguistic barriers. Developing techniques to improve factual consistency and reliability of generated summaries remains crucial, as does research into methods for handling diverse document formats and styles effectively. Additionally, exploring ways to enhance model scalability for processing very long or numerous documents efficiently will be vital for practical applications.

By addressing these challenges and further improving methodologies, long-context LLMs have the potential to become even more powerful tools for revolutionizing information processing and knowledge sharing in different fields. As research in this field progresses, it is crucial to focus on both improving technical capabilities and addressing ethical considerations. This balanced approach will be key to the effective and responsible use of long-context LLMs in multi-document summarization.

In conclusion, while long-context LLMs show immense promise, their successful application requires continued exploration and advancement. By tackling the identified obstacles and refining current approaches, we can work towards realizing the full potential of these models in transforming how we process and understand large volumes of information across diverse sectors and enterprise applications. The future of multi-document summarization with long-context LLMs is promising, but it demands rigorous research, ethical consideration, and innovative problem-solving to fully harness its capabilities.